\documentclass{article}

\usepackage{arxiv}

\usepackage[utf8]{inputenc} 
\usepackage[T1]{fontenc}    
\usepackage{hyperref}       
\usepackage{url}            
\usepackage{booktabs}       
\usepackage{amsfonts}       
\usepackage{nicefrac}       
\usepackage{microtype}      
\usepackage{lipsum}		
\usepackage{graphicx}
\usepackage{natbib}
\usepackage{doi}

\usepackage{caption}
\usepackage{subcaption}
\usepackage{cleveref}
\usepackage[table]{xcolor}
\usepackage{makecell}
\usepackage{multirow}

\usepackage{amssymb}
\usepackage{bbding}
\usepackage{pifont}
\usepackage{authblk} 
\usepackage{fancyhdr} 
\usepackage{hyperref} 
\title{RS-vHeat: Heat Conduction Guided Efficient Remote Sensing Foundation Model}

\date{} 					
\author[1,2,3,4]{Huiyang Hu}
\author[1,2,3,4]{Peijin Wang}
\author[1,2,3,4]{Hanbo Bi}
\author[1,2,3,4]{Boyuan Tong}
\author[2,5]{Zhaozhi Wang}
\author[1,2,3,4]{Wenhui Diao}
\author[1,2,3,4]{Hao Chang}
\author[1,2,3,4]{Yingchao Feng}
\author[3]{Ziqi Zhang}
\author[5]{Yaowei Wang}
\author[3,5]{Qixiang Ye}
\author[1,2,3,4]{Kun Fu}
\author[1,2,3,4]{Xian Sun}

\affil[1]{Aerospace Information Research Institute, Chinese Academy of Sciences}
\affil[2]{School of Electronic, Electrical and Communication Engineering, University of Chinese Academy of Sciences}
\affil[3]{University of Chinese Academy of Sciences}
\affil[4]{Key Laboratory of Target Cognition and Application Technology (TCAT)}
\affil[5]{Peng Cheng Laboratory}

\renewcommand{\shorttitle}{\textit{arXiv} Template}

\hypersetup{
pdftitle={A template for the arxiv style},
pdfsubject={q-bio.NC, q-bio.QM},
pdfauthor={David S.~Hippocampus, Elias D.~Striatum},
pdfkeywords={First keyword, Second keyword, More},
}

\begin{document}
\maketitle

\begin{abstract}
Remote sensing foundation models largely break away from the traditional paradigm of designing task-specific models, offering greater scalability across multiple tasks. However, they face challenges such as low computational efficiency and limited interpretability, especially when dealing with large-scale remote sensing images. To overcome these, we draw inspiration from heat conduction, a physical process modeling local heat diffusion. Building on this idea, we are the first to explore the potential of using the parallel computing model of heat conduction to simulate the local region correlations in high-resolution remote sensing images, and introduce RS-vHeat, an efficient multi-modal remote sensing foundation model. Specifically, RS-vHeat 1) applies the Heat Conduction Operator (HCO) with a complexity of $O(N^{1.5})$ and a global receptive field, reducing computational overhead while capturing remote sensing object structure information to guide heat diffusion; 2) learns the frequency distribution representations of various scenes through a self-supervised strategy based on frequency domain hierarchical masking and multi-domain reconstruction; 3) significantly improves efficiency and performance  over state-of-the-art techniques across 4 tasks and 10 datasets. Compared to attention-based remote sensing foundation models, we reduce memory usage by 84\%, FLOPs by 24\% and improves throughput by 2.7 times. The code will be made publicly available.
\end{abstract}


\section{Introduction}
Recently, remote sensing (RS) technology has become a vital data source for scientific research, resource management, and environmental monitoring \cite{sherrah2016fully, zhang2021cross, sun2022fair1m, li2020object} by capturing surface information via satellites. Traditional models, designed as single-task networks for specific RS tasks \cite{lu2021lil,sun2021pbnet}, struggle with multi-payload, multi-resolution, multi-temporal, and multi-feature RS data \cite{Kun2021Multi-satellite}. However, the emergence of remote sensing foundation models (RSFMs) has overcome these limitations, enabling unified handling of multiple tasks and diverse scenarios, significantly enhancing the scalability and versatility of the models \cite{li2021geographical, manas2021seasonal, mall2023change, ayush2021geography, cong2022satmae, wang2022advancing, tao2023tov, reed2023scale, mendieta2023towards, bastani2023satlaspretrain}. By constructing visual encoders, RSFMs can automatically extract and learn features from remote sensing imagery (RSI), providing a robust foundation for various real-world RS tasks.

\begin{figure}[tbp]
    \centering
    \includegraphics[width=0.5\linewidth]{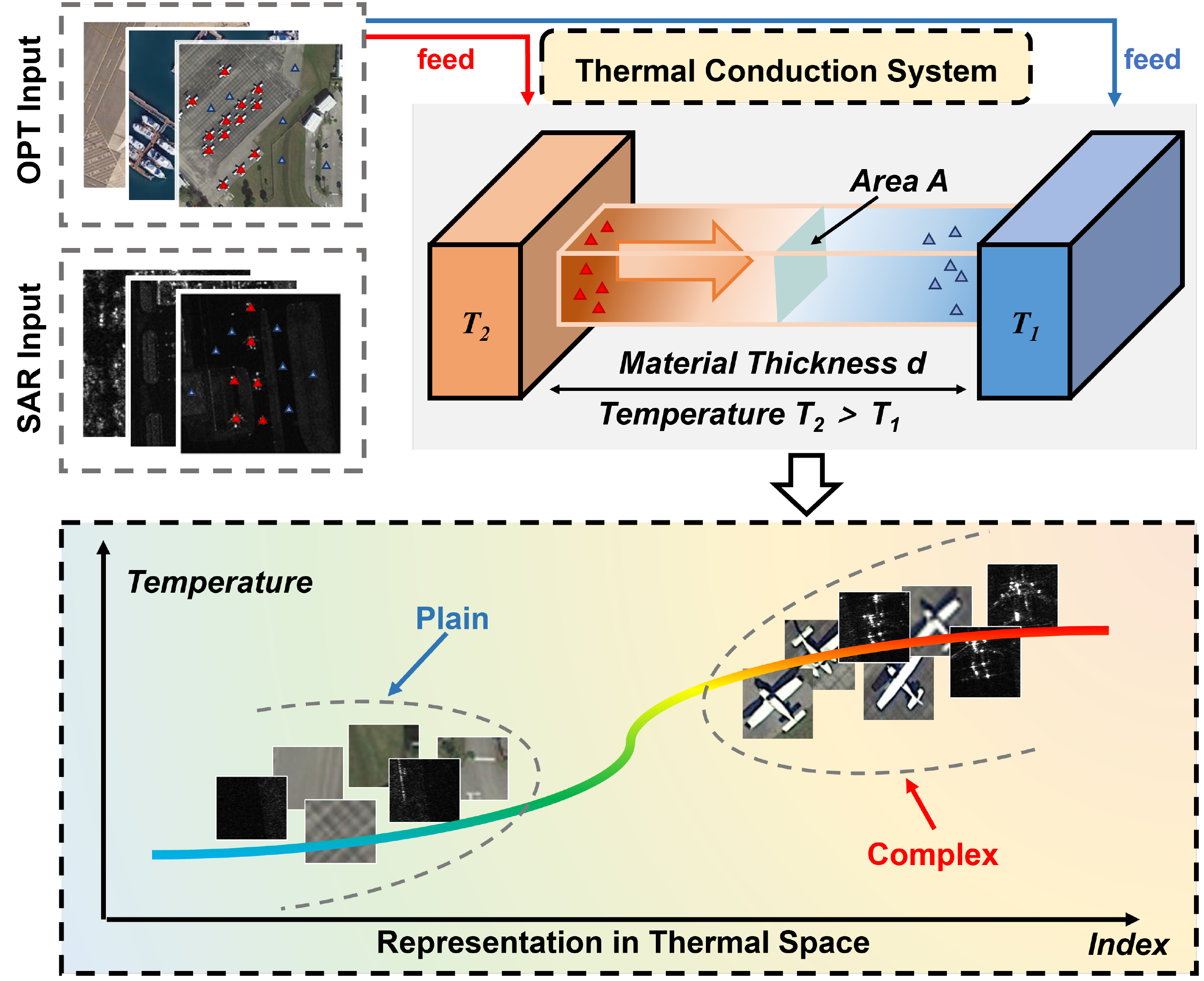}
\caption{Calculation of RS images in the heat conduction system. In heat conduction theory, different materials exhibit varying diffusion rates. Inspired by this, the heat-conduction-based encoder maps  optical (OPT) and synthetic aperture radar (SAR) into a unified thermal space, enhancing sensitivity to shape features by simulating heat flow and accumulation within the object region.}
\label{fig:aaa_rm_heat_conduction}

\end{figure}  

Prior RSFMs typically use visual encoders and decoders, leveraging large-scale RS datasets for pre-training while incorporating self-supervised learning strategies. Specifically, RSFMs can be categorized based on the type of backbone used: CNN-based methods often use ResNet18/50 \cite{he2016deep} to extract features \cite{li2021geographical,manas2021seasonal,mall2023change,ayush2021geography}, while attention-based methods mainly use ViT \cite{dosovitskiy2020image} or Swin transformer \cite{liu2021swin} with attention mechanisms \cite{vaswani2017attention} to capture global dependencies \cite{cong2022satmae,wang2022advancing,tao2023tov,reed2023scale,mendieta2023towards,bastani2023satlaspretrain}. Both approaches use strategies such as masked reconstruction, knowledge distillation, or contrastive learning during pre-training to enhance model robustness (a detailed comparison is available in the supplementary materials). Despite the significant advancements achieved by these RSFMs, they still face two limitations:

\noindent\textbf{Balancing Efficiency and Receptive Field.} To accurately capture information about the large objects in RSI, model outputs must be responsive to sufficiently large regions \cite{NIPS2016_c8067ad1}. However, this need significantly increases computational complexity \cite{christophe2011remote, MA201547}. CNN-based networks lack a global receptive field due to their  reliance on sliding computations with fixed-size convolutional kernels. While attention-based models achieve global modeling, their attention mechanisms incur quadratic computational complexity. Therefore, existing RSFMs struggle to deliver both fast and high-accuracy inference in practical applications.

\noindent \textbf{Weak Physical Interpretability.} RS objects often exhibit irregular polygonal shapes \cite{li2003shape}, and current RSFMs struggle to integrate physical principles to explain how object features propagate. This deficiency makes it challenging for researchers to adjust learning strategies effectively \cite{9829013, 10057399, 9323898}. In the long term, RSFMs need to possess a certain degree of information interpretability.

\begin{figure*}[htbp]
    \centering
    \includegraphics[width=\linewidth]{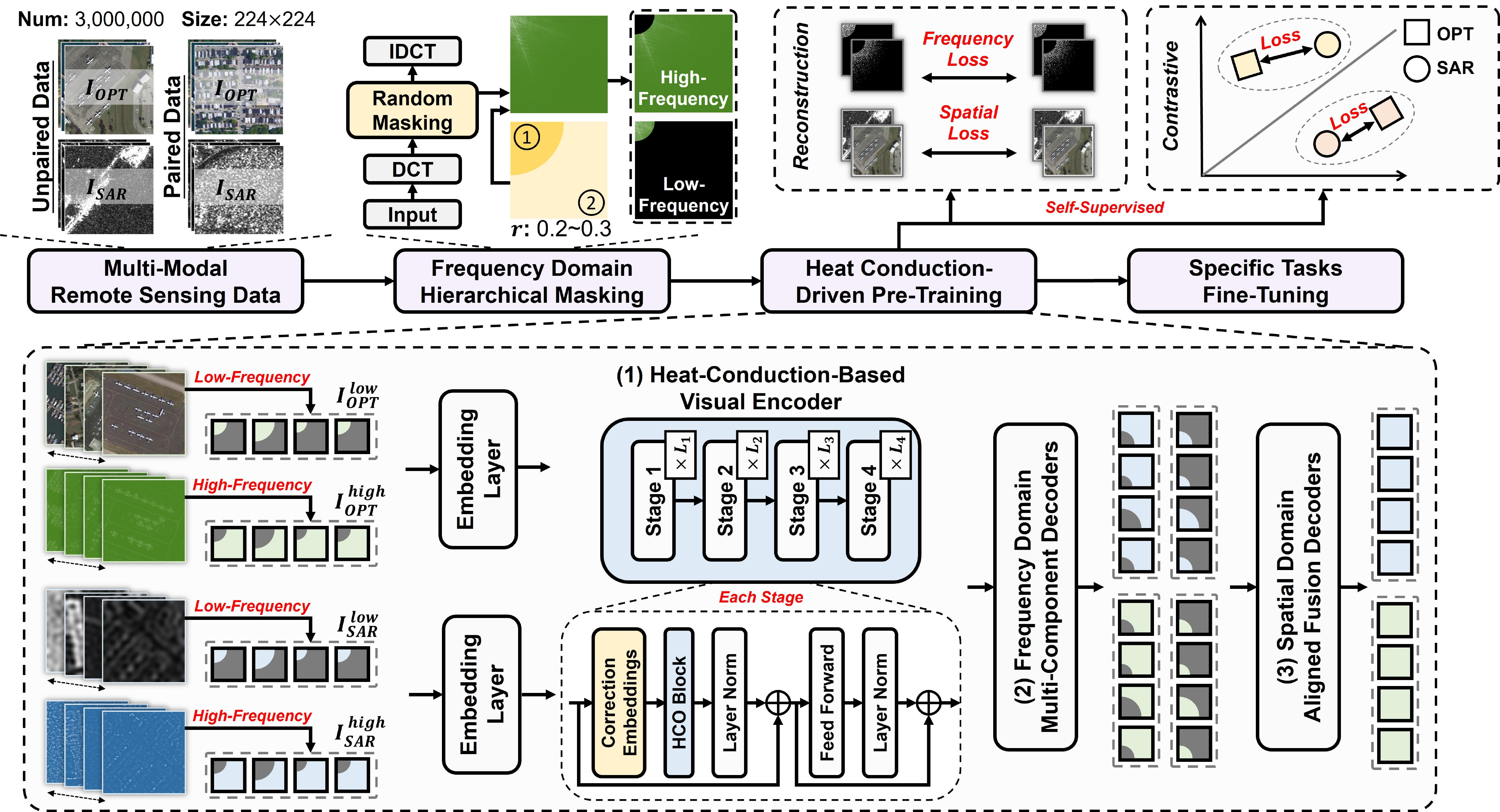}
    \caption{ The pre-training process of RS-vHeat. It performs frequency domain hierarchical masking based on randomly generated sector regions to separate each image into high- and low-frequency components. These component images are fed into the network and projected into the thermal space via embedding layers. The heat diffusion is computed within the heat-conduction-based visual encoder to simulate complex RS objects. The deep features undergo multi-domain reconstruction loss and contrastive loss computation via decoders. }
\label{fig:aaa_rs-vheat_architecture}

\end{figure*}

To address these, this paper introduces RS-vHeat, a heat-conduction-based RSFM that supports multi-modal inputs inspired by the idea of vHeat \cite{wang2024vheat}. \textbf{First}, heat conduction represents the natural process of energy diffusion from high- to low-temperature regions, as shown in the conceptual physical model in \cref{fig:aaa_rm_heat_conduction}. This process transitions from an unsteady to a steady state based on the material. Since its computational process resembles feature extraction in neural networks, it can be applied to RS image processing. \textbf{Second}, we hypothesize that object types correspond to special feature distributions, with the model predicting diffusion rates based on RS-specific properties, simulating parameter computation through heat flow. This approach projects all modalities into a common thermal space, following the constraint that complex areas containing RS objects are high-temperature regions where heat accumulates, while sparse regions are low-temperature areas where heat diffuses easily, as shown in the lower part of \cref{fig:aaa_rm_heat_conduction}. \textbf{Third}, building on this theory, we further design a RSFM with physical interpretability using 3 million optical and SAR data for pre-training, as shown in \cref{fig:aaa_rs-vheat_architecture}. The heat conduction network simulates the diffusion process of heat across large-scale multi-modal RS data, facilitating the alignment of feature propagation with the structural characteristics of the objects. Moreover, its computational approach provides guidance for the efficient operation of the network.

To summarize, our contributions are as follows:

\begin{enumerate}
    \item We introduce RS-vHeat, a RSFM designed based on the heat conduction differential equation to process RS data. It conceptualizes the semantic relationships between pixels in RS images as the propagation of heat.
    \item We propose a self-supervised strategy based on frequency domain hierarchical masking and multi-domain reconstruction that preserves small objects, driving the model to reconstruct fine and coarse frequency signals. 
    \item We design spatial correction embeddings, which operate directly on the global features to capture local details, assisting in simulating the rate of thermal diffusion.
    \item We evaluate RS-vHeat on 10 datasets, showing it outperforms advanced RSFMs in accuracy while maintaining lower computational complexity. When processing large-scale images, RS-vHeat reduces memory usage by 84\%, decreases FLOPs by 24\%, and improves throughput by 2.7 times compared to attention-based RSFMs.
\end{enumerate}

\section{Related Work}

\subsection{Mainstream Network Architectures in RS}

\noindent\textbf{Convolutional Neural Networks.} CNNs have been adapted for RS to tackle challenges in satellite imagery \cite{zhang2019detecting,li2020adversarial}. However, local receptive fields, mainly limited by the kernel size, restrict the capture of long-range dependencies \cite{dosovitskiy2020image}. This drawback is especially critical in RS, where large-scale patterns and complex spatial relationships are prevalent, making the balance between local feature extraction and global context a key research focus \cite{dong2021remote,chen2020adaptive}.

\noindent\textbf{Transformers.} Self-attention mechanisms \cite{vaswani2017attention} empower networks to capture long-range dependencies \cite{dosovitskiy2020image,liu2021swin}. Research studies have shown that plain ViTs performs better than traditional CNN models in RS applications \cite{wang2022advancing,yan2022fully}. However, the limitation stems from the ViTs' restricted ability to effectively handle long sequences. When processing large-scale RS images, the computational load of networks grows quadratically, leading to significant overhead. 

\noindent\textbf{Mamba-Based Models.} Mamba, an efficient implementation of State Space Models (SSMs) \cite{gu2021efficiently}, leverages selective scan mechanisms for long-sequence processing \cite{liu2024vmambavisualstatespace,zhu2024vision}, balancing computational efficiency and high accuracy. \cite{gu2023mamba, xu2024survey}. It has been rapidly adopted in RS \cite{chen2024rsmamba,zhu2024samba,chen2024changemamba}, with its ability to process long sequences, although Mamba, as an innovative architecture, suffers from limited interpretability.

\subsection{Self-Supervised Learning Strategies for RSFMs}
\noindent\textbf{Contrastive Learning.} By establishing rules to distinguish between positive and negative samples, contrastive learning aims to fully understand the relationships between these samples. GASSL \cite{ayush2021geography} treats RS images of the same scene captured at different times as positive pairs, using them as self-supervised signals. Based on seasonal contrast method, SeCo \cite{manas2021seasonal} and CACo \cite{mall2023change} effectively leverage temporal information within the networks by comparing scene variations across different years or seasons as perceptual signals.  Skysense \cite{guo2024skysense} utilizes multi-modal RS images as input and implements a multi-granularity contrastive learning framework. Motivated by the aforementioned methods, we apply contrastive constraints in the thermal space to optical and SAR data, encouraging the model to focus on deep, fine-grained semantic relationships within the RS images.

\noindent\textbf{Masked Image Modeling.} By involving masking parts of the image and predicting the missing information, networks can understand the details of RSI. RingMo \cite{sun2022ringmo} attempts to solve the issue that directly masking image patches can easily lead to the loss of small objects by designing an incomplete masking strategy that is implemented proportionally within the patches. SpectralGPT \cite{hong2024spectralgpt} models multi-spectral data as three-dimensional data and applies masking in three-dimensional space. Scale-MAE \cite{reed2023scale} masks pixels in the spatial domain and reconstructs high- and low-frequency images to learn image representations at different scales. Inspired by these, to preserve and learn the information of complex RS scenes, we employ dual reconstruction in both the spatial and frequency domains as constraints.

\section{Proposed RS-vHeat}
We introduce RS-vHeat, a multi-modal remote sensing foundation model with three key components: a frequency domain hierarchical masking strategy for multi-modal RS data, a visual encoder for modeling internal heat flow within RS images, and decoders for multi-domain reconstruction. \cref{fig:aaa_rs-vheat_architecture} illustrates the RS-vHeat structure during pre-training, enabling shared representation of multi-modal data through frequency domain masking and projection into a thermal space to simulate heat propagation. Preliminary details on vHeat \cite{wang2024vheat} are included in the supplementary materials.

\begin{figure*}[tbp]
    \centering
    \includegraphics[width=\linewidth]{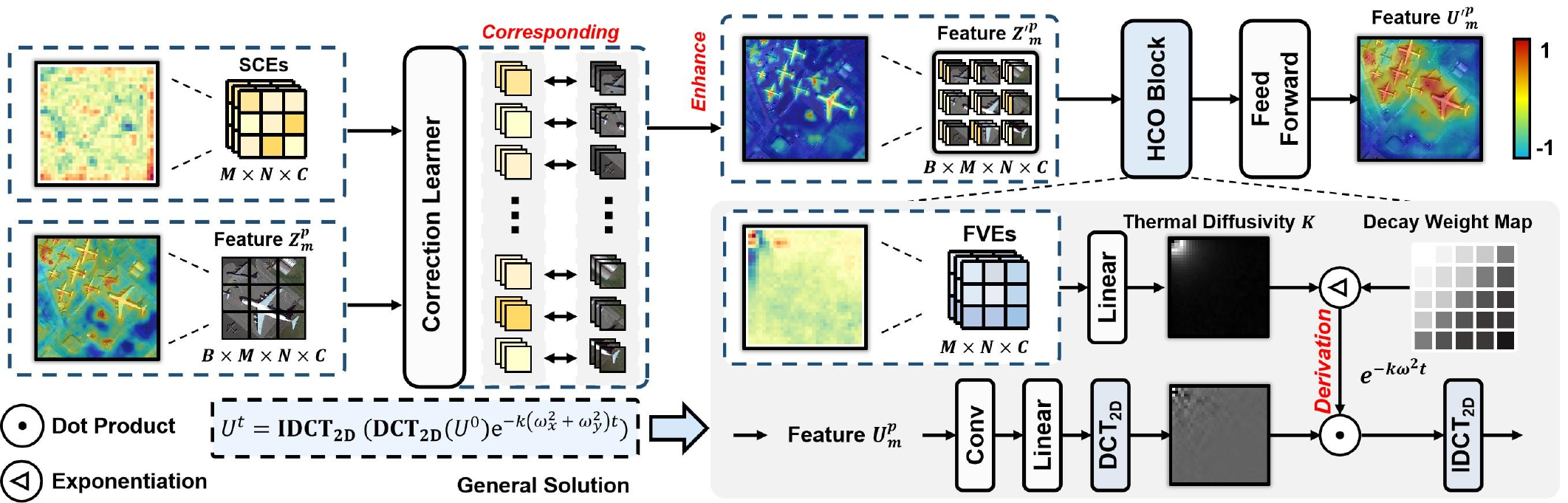}
    \caption{The overall structure of the heat-conduction-based visual encoder, simulates the general solution of heat conduction. SCEs first dynamically perform spatial domain correction, adjusting heat distribution based on the RS scene. The enhanced image undergoes a 2D DCT in the HCO block, interacting with the heat diffusion rate predicted by the FVEs to complete a heat conduction calculation, which is then transformed back into a visual representation via 2D IDCT.}
\label{fig:aaa_feature_vis}

\end{figure*} 

\noindent\textbf{Frequency Domain Hierarchical Masking Strategy.} Traditional spatial masking \cite{xie2022simmim} on RS images often obscures smaller objects, hindering reconstruction. Inspired by the frequency-aware dynamic network \cite{xie2021learning}, we implement a frequency domain hierarchical masking strategy to preserve object structure, enabling more accurate heat propagation by focusing on frequency domain signals.

Specifically, given the input (optical and SAR), denoted as \( I(x,y,c)=\{I_{o}, I_{s}\}, I_{o} \in \mathbb{R}^{\mathit{H \times W \times} 3}, I_{s} \in \mathbb{R}^{\mathit{H \times W \times} 1} \), two parallel streams process paired and unpaired data. The Discrete Cosine Transform (DCT) is applied along each image dimension, converting  \( I(x, y) \) to its frequency representation $\tilde{I}(u,v)$, where low-frequency components are concentrated in the top-left of the spectrum. A sector mask separates high-frequency \( \tilde{I}^{high}(u,v) \) and low-frequency \( \tilde{I}^{low}(u,v) \) regions, using a random masking rate of 20\%-30\%. Next, the Inverse Discrete Cosine Transform (IDCT) is applied along the image dimensions to revert the data back to the spatial domain, obtaining \( I^{low}(x,y) \) and \( I^{high}(x,y) \). Notably, the DCT and IDCT operations are differentiable, enabling efficient computation and seamless integration into network training on both CPUs and GPUs. Details can be found in the supplementary material.

\noindent\textbf{Heat-Conduction-Based Visual Encoder.} High- and low-frequency information from multi-modal data is mapped to a shared thermal space and then fed into the heat-conduction-based visual encoder for thermal simulation. In most RS downstream tasks, areas containing RS objects tend to appear as high-temperature zones, while sparse or empty regions display as low-temperature zones. 

The Heat Conduction Operator (HCO) simulates the process of visual information transmission as thermal conduction. The two-dimensional temperature distribution at time t, $u(x, y, t)$ is extended to a multi-dimensional feature distribution $U(x, y, c, t)$. The HCO block specifically models the general solution of physical heat conduction:
\begin{equation}
U_m^t = \mathcal{F}^{-1}\Big(\mathcal{F}(U_m^0)e^{-k(\omega^2_x +\omega^2_y)t}\Big)
\end{equation}
where $U^0$ and $U^t$ represents the input $U(x, y, c, 0)$ and output $U(x, y, c, t)$. We denote the Discrete Fourier Transform (DFT) and its inverse (IDFT) as $\mathcal{F}$ and $\mathcal{F}^{-1}$. \( m \in \{o, s\} \) refers to the modality (optical or SAR). Based on the Neumann boundary condition \cite{cheng2005heritage}, we replace the 2D DFT and IDFT with the 2D DCT (DCT\textsubscript{2D}) and IDCT (IDCT\textsubscript{2D}) \cite{strang1999discrete}:
\begin{equation}
U_m^t = IDCT_{2D}\Big(DCT_{2D}(U_m^0)e^{-k(\omega^2_x +\omega^2_y)t}\Big)
\end{equation}
where \( e^{-k(\omega^2_x + \omega^2_y)t} \) functions as an adaptive filter in the frequency domain, executing heat conduction.

\begin{figure*}[tbp]
    \centering
    \includegraphics[width=\linewidth]{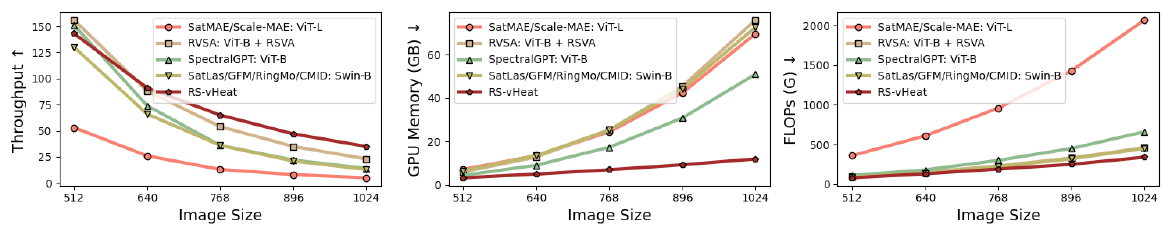}
    \caption{ Comparison of throughput (left), GPU memory (middle), and FLOPs (right) across image sizes for RSFMs. RS-vHeat shows significantly higher throughput, lower memory usage and lower FLOPs than current attention-based RSFMs, especially for large-scale images. All tests were performed in a consistent environment using a single A100 (80G) GPU with a batch size of 32.}
\label{fig:memory_vs_image_size}
\end{figure*} 

A set of learnable frequency value embeddings (FVEs) is predicted to estimate the heat diffusion coefficient $k$ $(k:=k(\omega_x, \omega_y))$, facilitating adaptive heat transfer, as shown in \cref{fig:aaa_feature_vis}. The weights are denoted as \( W_{FVEs} \in \mathbb{R}^{M \times N \times C} \):
\begin{equation}
U^{\prime p}_{m}= HCO(U^{p}_{m}, W_{FVEs})
\end{equation}
where \( U \) and \( U\prime  \) represent the temperature states of the features from different modalities before and after passing through the HCO block, respectively, with \( p \in \{ high, low\} \) indicating the frequency level. 

To enhance the spatial representation of different structural components, we introduce a lightweight correction learner. By predicting a set of spatial correction embeddings (SCEs) from large-scale pre-trained images, it interacts with the existing temperature field and perform activation, as shown in \cref{fig:aaa_feature_vis}. The weight \( W_{SCEs} \in \mathbb{R}^{\mathit{M \times N \times C}} \) adaptively adjusts the object boundaries in the spatial domain based on content by adding them to the original image \( I(x, y) \).  Given the upper-layer temperature feature \( Z_{m}^{p} \), the correction learner can be expressed as \( CL \):
\begin{equation}
Z^{\prime p}_{m}= CL(Z^{p}_{m}, W_{SCEs})
\end{equation}
Where \( Z^{\prime p}_{m} \) represents the temperature features after adaptive adjustment. 

\noindent\textbf{Multi-Domain Reconstruction Decoders.}
Considering that the restoration of heat distribution is modality-specific, the optical and SAR decoders (\cref{fig:aaa_rs-vheat_architecture} (2)), \(D^{\delta}_{o}\) and \(D^{\delta}_{s}\) concurrently operate on the encoded outputs (\(F^{\prime p}_{o} \), \(F^{\prime p}_{s} \in \mathbb{R}^{M \times N \times C}\)). By employing convolution and pixel shuffling, decoders effectively upsample feature maps. The reconstruction outputs \(I^{\prime }\) for both modalities—\(I^{\prime high}_{o}\), \(I^{\prime low}_{o}\) for optical, and \(I^{\prime high}_{s}\), \(I^{\prime low}_{s}\) for SAR—are as follows:
\begin{equation}
I^{\prime p}_{o} = D^{\delta}_{o}(F^{\prime p}_{o}),I^{\prime p}_{s} = D^{\delta}_{o}(F^{\prime p}_{s})
\end{equation}

The reconstruction is guided by the \(\mathcal{L}_1\) loss in the frequency domain. After applying DCT,  the loss  \( \mathcal{L}_{Fre} \) is computed by measuring the difference between transformed features \( \tilde{I}^{\prime p}_m(u,v) \) and \( \tilde{I}^p_m(u,v) \). 

To learn more robust higher-level features, spatial decoders (\cref{fig:aaa_rs-vheat_architecture} (3)) are responsible for fusion and reconstruction. Specifically, outputs,  \({F}^{\prime p}_{m}\) and \({F}^{p}_{m}\), from the third and fourth stages extract modality-specific frequency features. The fusion layer integrates the frequency components into higher-level image features, \({F}^{\prime }_{m}\) and \({F}_{m}\):
\begin{equation}
F^{\prime }_{m} = Concat[CONV(F^{\prime high}_{m}),CONV(F^{\prime low}_{m})], 
F_{m} = Concat[CONV(F^{high}_{m}),CONV(F^{low}_{m})] \\
\end{equation}

Furthermore, the integration of multi-stage features is achieved through the function  \( P_{m} = g(F_{m}, {F}^{\prime }_{m}) \), where \(g\) combines convolution, ReLU activation and element-wise summation operations to collectively produce the final high-level output \(P_{m} = \{P_{o},P_{s}\}\). Subsequently, these features are fed into modality-specific spatial domain decoders \(D^{\phi}_{o}\) and \(D^{\phi}_{s}\), they are upsampled to the original image size to restore the high-level semantics of the original image:
\begin{equation}
I^{\prime }_{o} = D^{\phi}_{o}\big(P_{o}\big), I^{'}_{s} = D^{\phi}_{s}\big(P_{s}\big)
\end{equation}
Where \( I'_{o} \) and \( I'_{s} \) represent the reconstructed high-level optical and SAR data. The reconstruction loss in the spatial domain \( \mathcal{L}_{Spa} \)  is then computed using the \(\mathcal{L}_1\) loss.

To explore the network's understanding of fine-grained semantic information, we employ contrastive loss to compute the semantic discrepancy between different modalities. The loss operates on embeddings $F_{m}^{p}$ obtained from two different preprocessing methods applied to the same image, representing low- and high-frequency features. The contrastive loss \(\mathcal{L}_{Con}\) is calculated leveraging cosine similarity, following the computational methodology outlined in \cite{atito2021sit}.

The overall loss function is shown in \cref{loss}. The combined approach encourages the model to fully capture both global structures and fine-grained details:
\begin{equation}
\mathcal{L}_{total} = \mathcal{L}_{Con} + \mathcal{L}_{Spa} +\mathcal{L}_{Fre}
\label{loss}
\end{equation}

\section{Experiments}

\subsection{Training Implementation}

The training of RS-vHeat consists of self-supervised pre-training and downstream fine-tuning. The visual encoder follows the Swin-B \cite{liu2021swin} configuration with four stages of 2, 2, 18, and 2 blocks. During pre-training, we use a large-scale multi-modal dataset with 450k matched optical and SAR image pairs, totaling over 3 million entries, following the methodology of RingMo \cite{sun2022ringmo}. The model is trained on eight A100 (80G) GPUs for 200 epochs with images of size 224. The training starts with a learning rate of 1e-6 for 10 warm-up epochs, gradually increasing to 2e-4, with a cosine annealing schedule and a minimum learning rate of 1e-5. For fine-tuning, we transfer the pre-trained embedding layers and visual encoder structure, adjusting the fixed-size FVEs and SCEs to match the image dimensions.

\subsection{Performance}

We comprehensively evaluate the performance of RS-vHeat compared to other representative RSFMs. Primarily, as shown in \cref{fig:memory_vs_image_size}, we analyze throughput, memory usage, and FLOPs across various image sizes. RS-vHeat (depicted by the red line) demonstrates significantly higher throughput, lower memory consumption and FLOPs, with these advantages becoming more pronounced as image sizes increase. For instance, when processing $1024\times1024$ resolution images, RS-vHeat achieves 2.7 times the throughput of Swin-B-based models such as SatLas \cite{bastani2023satlaspretrain}, GFM \cite{mendieta2023towards}, RingMo \cite{sun2022ringmo}, and CMID \cite{muhtar2023cmid}, while reducing memory usage by 84\%. Similarly, when compared to the improved ViT-B-based SpectralGPT \cite{hong2024spectralgpt}, RS-vHeat achieves 2.5 times the throughput while reducing memory usage by 77\%.

Next, we evaluate the accuracy and computational efficiency of fine-tuning RS-vHeat on four tasks—semantic segmentation, object detection, classification and change detection—to demonstrate the effectiveness, along with extensive ablation studies. Many RSFMs primarily focus on optical data and do not support SAR input, so we compare RS-vHeat with other RSFMs using optical datasets, while SAR-specific tasks are compared against specialized models. RSFMs are categorized into three main groups based on the backbone: CNN-based, ViT-based, and Swin-based. All datasets follow official partitioning methods for training and testing. More details on datasets and visualizations are provided in the supplementary materials.

\noindent\textbf{Single- and Multi-Modal Semantic Segmentation.} We evaluate our model on two optical datasets (Potsdam \cite{sherrah2016fully}, iSAID \cite{waqas2019isaid}), one SAR dataset (Air-PolSAR-Seg\cite{wang2022air}), and one multi-modal dataset (\cite{li2022mcanet}), using UPerNet \cite{Xiao_2018_ECCV} with cross-entropy loss for the output head. \cref{tab:isaid_potsdam} compares RS-vHeat with 14 other RSFMs on iSAID and Potsdam. RS-vHeat demonstrates superior accuracy compared to CNN-based models, and achieves lower FLOPs than ViT-based models, exemplified by its 1.28\% and 2.95\% improvement over Scale-MAE \cite{reed2023scale}. It also exhibits the lowest FLOPs among Swin-based models, with over a fourfold reduction in parameter count compared to SkySense \cite{guo2024skysense}, with only 1.17\% and 2.19\% decreases in accuracy. Besides, on AIR-PolSAR-Seg (\cref{tab:air-polsar-seg}) and WHU-OPT-SAR (\cref{tab:whu-opt-sar}), RS-vHeat outperforms other specialized segmentation models, demonstrating the applicability of the heat conduction model in SAR and multi-modal segmentation tasks.

\begin{table*}[htbp]
\centering
\setlength{\tabcolsep}{12pt}  
\caption{ Comparison of mF1, mIoU, parameters, and FLOPs on iSAID and Potsdam with other RSFMs.}
\resizebox{1.0\linewidth}{!}{
\begin{tabular}{c|r|c|c|c|c|c|c}
\toprule
Backbone Type & Method  & Backbone  & \makecell[c]{\underline{Potsdam} \\ mF1 $\uparrow$ }  & \makecell[c]{\underline{iSAID} \\ mIoU $\uparrow$} &Image Size   & Params  & FLOPs $\downarrow$  \\
\midrule
\multirow{6}{*}{CNN-Based}
&GASSL (ICCV’2021) \cite{ayush2021geography} & ResNet-50   & 91.27 & 65.95 & $896^{2}$ & 64M & 722G\\
&SeCo (ICCV'2021) \cite{manas2021seasonal} & ResNet-50 & 89.03  & 57.20  & $896^{2}$ & 64M & 722G\\
&SSL4EO (GRSM'2023) \cite{wang2023ssl4eo}& ResNet-50  & 91.54  & 64.01& $896^{2}$ & 64M & 722G \\
&CACo (CVPR’2023) \cite{mall2023change} & ResNet-18   & 91.35 & 64.32 & $896^{2}$ & 41M & 671G\\
&TOV (JSTARS’2023) \cite{tao2023tov}& ResNet-50   & 92.03  & 66.24 & $896^{2}$ & 64M & 722G\\
&SAMRS (NeurIPS'2023) \cite{wang2024samrs} & ResNet-50 & 91.43 & 66.26  & $896^{2}$ & 64M & 722G\\
 & \cellcolor{gray!10}RS-vHeat (Ours) &  \cellcolor{gray!10}vHeat-B + SCEs &  \cellcolor{gray!10}\textbf{92.82}  &  \cellcolor{gray!10}\textbf{68.72}   &  \cellcolor{gray!10}$896^{2}$ &  \cellcolor{gray!10}148M & \cellcolor{gray!10}921G \\
\midrule
\multirow{5}{*}{ViT-Based}
&RVSA (TGRS'2022) \cite{wang2022advancing} & ViT-B + RVSA  & -  & 64.49 & $896^{2}$ &128M &1043G\\
&SatMAE (NeurIPS'2022)\cite{cong2022satmae}& ViT-L & 90.63 & 62.97  & $896^{2}$ & 341M & 1536G  \\
&Scale-MAE (ICCV'2023) \cite{reed2023scale}& ViT-L   & 91.54 & 65.77 & $896^{2}$ & 341M & 1536G \\
 & \cellcolor{gray!10} RS-vHeat (Ours) & \cellcolor{gray!10}vHeat-B + SCEs  & \cellcolor{gray!10}\textbf{92.82} &  \cellcolor{gray!10}\textbf{68.72} & \cellcolor{gray!10}$896^{2}$ & \cellcolor{gray!10}148M & \cellcolor{gray!10}\textbf{921G} \\
\midrule
\multirow{6}{*}{Swin-Based}
&RingMo (TGRS'2022) \cite{sun2022ringmo}  & Swin-B & 91.27 & 67.20  & $896^{2}$  & 121M &968G\\
&SatLas (ICCV'2023) \cite{bastani2023satlaspretrain}& Swin-B & 91.28 &  68.71  & $896^{2}$ & 121M &968G\\
&GFM (ICCV'2023) \cite{mendieta2023towards}& Swin-B   & 91.85 & 66.62 & $896^{2}$ & 121M &968G\\
&CMID (TGRS'2023) \cite{muhtar2023cmid}& Swin-B & 91.86  & 66.21 & $896^{2}$ & 121M &968G\\
&SkySense (CVPR'2024) \cite{guo2024skysense} & Swin-H & \textbf{93.99} & \textbf{70.91} & $896^{2}$  & \textgreater 702M  & \textgreater 2708G \\
 & \cellcolor{gray!10} RS-vHeat (Ours) & \cellcolor{gray!10} vHeat-B + SCEs & \cellcolor{gray!10}92.82 & \cellcolor{gray!10}68.72 & \cellcolor{gray!10}$896^{2}$ & \cellcolor{gray!10}148M & \cellcolor{gray!10}\textbf{921G}\\
\bottomrule 
\end{tabular}
}
\label{tab:isaid_potsdam}
\vspace{-1.2em}
\end{table*}

\begin{table*}[htbp]
\centering
\setlength{\tabcolsep}{12pt}  
\caption{ Comparison of OA and user’s accuracy on WHU-OPT-SAR with other specialized models.}
\resizebox{1.0\linewidth}{!}{
\begin{tabular}{r|c|c|c|c|c|c|c|c|c}
\toprule

\multicolumn{1}{r|}{\multirow{2}{*}{Method}} & \multirow{2}{*}{Publication}
& \multicolumn{7}{c|}{User's Accuracy $\uparrow$} & \multicolumn{1}{r}{\multirow{2}{*}{OA $\uparrow$}} 
\\  \cline{3-9}
  & \multicolumn{1}{r|}{} & Farmland & City & Village & Water & Forest & Road & Others      
 &\\  
\midrule
SegFormer \cite{xie2021segformer}   &NeurIPS'2021            & 79.1          & 72.9          & 38.0          & 64.7          & 88.1          & 0.3           & 0.4           & 75.5\\

Segmenter \cite{strudel2021segmenter}     &ICCV'2021        & \textbf{82.3}         & \textbf{75.2}          & 51.7          & 74.4          & 89.4          & 16.0          & 12.0          & 79.9 \\
MCANet \cite{li2022mcanet}  &JAG'2022 &74.3	&62.2	&53.1	&65.7	&\textbf{95.5}	&31.0	&9.8 &82.9\\
VMamba \cite{liu2024vmambavisualstatespace}  &NeurIPS'2024 & 82.1 & 74.1 & 57.8 & \textbf{81.4} & 89.1 & 41.5 & 18.4 & 81.4 \\
MMOKD \cite{liu2024multimodal} &TGRS’2024 & 70.0 & 58.1 & 50.0 & 69.7 & 80.1 & 40.1 & 25.0 & 82.5
 \\
\midrule
\cellcolor{gray!10} RS-vHeat (Ours) & \cellcolor{gray!10}- &  \cellcolor{gray!10}81.1 & \cellcolor{gray!10}67.3 &  \cellcolor{gray!10}\textbf{67.5}& \cellcolor{gray!10}79.0 & \cellcolor{gray!10} 90.2 & \cellcolor{gray!10} \textbf{54.9} & \cellcolor{gray!10} \textbf{55.3} &\cellcolor{gray!10}\textbf{83.9} \\
\bottomrule
\end{tabular}
}
\label{tab:whu-opt-sar}
\vspace{-1.2em}
\end{table*}

\begin{table}[htbp]
\centering
\setlength{\tabcolsep}{6pt}  
\caption{Comparison of mIoU, OA, and AA on AIR-PolSAR-Seg with other specialized models.}
\resizebox{0.5\linewidth}{!}{
\begin{tabular}{r|c|c|c|c}
\toprule
Method & Publication & mIoU $\uparrow$ & OA $\uparrow$ & AA $\uparrow$ \\
\midrule
DeepLab V3+ \cite{chen2018encoder}  & ECCV'2018  & 48.21 & 76.81 & 63.55 \\
EncNet \cite{zhang2018context} & CVPR'2018   & 47.75 & 75.67 & 57.51 \\
PSANet \cite{zhao2018psanet}  & ECCV'2018  & 47.14 & 76.21 & 62.92 \\
CCNet \cite{huang2019ccnet} &  ICCV'2019  & 46.46 & 75.53 & 55.83 \\
DANet \cite{fu2019dual}  &  CVPR'2019 & 51.93 & 76.91 & 62.79 \\
GCNet \cite{cao2019gcnet}  & ICCV'2019  & 47.56 & 76.75 & 57.10 \\ \cmidrule(lr){1-5}
\cellcolor{gray!10} RS-vHeat (Ours) &   \cellcolor{gray!10}- & \cellcolor{gray!10}\textbf{57.46} & \cellcolor{gray!10}\textbf{81.46} &\cellcolor{gray!10} \textbf{65.92} \\
\bottomrule
\end{tabular}
\label{tab:air-polsar-seg}
}
\end{table}

\noindent\textbf{Object Detection.} We conduct coarse- and fine-grained experiments on two optical datasets (FAIR1M \cite{sun2022fair1m}, DIOR \cite{li2020object}) and one SAR dataset (SAR-AIRcraft-1.0 \cite{zhirui2023sar}), using YOLOX  \cite{ge2021yolox} as the output head. On the DIOR dataset, \cref{tab:dior} shows that RS-vHeat demonstrates not only lower FLOPs but also surpasses the results of SkySense \cite{guo2024skysense} by 3.57\%. As shown in \cref{tab:object_detection_SAR-AIRcraft} and  \cref{tab:FAIR1M-2.0}, RS-vHeat also achieves strong performance and high computational efficiency on two additional fine-grained datasets.

\begin{table*}[htbp]
\caption{Comparison of mAP\textsubscript{50}, parameters, and FLOPs on DIOR with other RSFMs.}
\setlength{\tabcolsep}{12pt}  
\resizebox{1.0\textwidth}{!}{
\begin{tabular}{c|r|c|c|c|c|c}
\toprule
 Backbone Type & Method &  Backbone &  Image Size &  mAP\textsubscript{50}$\uparrow$  &  Params &  FLOPs $\downarrow$ \\
\midrule
\multirow{4}{*}{CNN-Based}
&GASSL (ICCV’2021) \cite{ayush2021geography} & ResNet-50& $800^{2}$& 67.40  & 41M & 134G\\
& CACo (CVPR’2023) \cite{mall2023change} & ResNet-18  & $800^{2}$ & 66.91& 28M & 101G\\

&TOV (JSTARS’2023) \cite{tao2023tov}& ResNet-50 & $800^{2}$& 70.16   & 41M & 134G\\
&SSL4EO (GRSM'2023) \cite{wang2023ssl4eo}& ResNet-50  & $800^{2}$& 64.82 &41M & 134G\\

& \cellcolor{gray!10}RS-vHeat (Ours)& \cellcolor{gray!10} vHeat-B + SCEs & \cellcolor{gray!10} $800^{2}$ & \cellcolor{gray!10}\textbf{82.30}&  \cellcolor{gray!10}128M &  \cellcolor{gray!10}266G \\
\midrule

\multirow{5}{*}{ViT-Based}
&RVSA (TGRS'2022) \cite{wang2022advancing} & ViT-B + RVSA & $800^{2}$& 73.22  &113M &378G \\
&SatMAE (NeurIPS'2022)\cite{cong2022satmae}& ViT-L  & $800^{2}$& 70.89 &324M &1094G \\
&Scale-MAE (ICCV'2023) \cite{reed2023scale}& ViT-L  & $800^{2}$& 73.81 &324M &1094G \\ 
&  \cellcolor{gray!10}RS-vHeat (Ours) &  \cellcolor{gray!10}vHeat-B + SCEs &  \cellcolor{gray!10}$800^{2}$  &  \cellcolor{gray!10}\textbf{82.30}&  \cellcolor{gray!10}128M & \cellcolor{gray!10} \textbf{266G}\\
\midrule
\multirow{6}{*}{Swin-Based} 
&RingMo (TGRS'2022) \cite{sun2022ringmo}  & Swin-B& $800^{2}$& 75.90 &105M & 322G\\
&SatLas (ICCV'2023) \cite{bastani2023satlaspretrain}& Swin-B& $800^{2}$ & 74.10  &105M & 322G\\
&GFM (ICCV'2023) \cite{mendieta2023towards}& Swin-B & $800^{2}$ & 72.84 &105M & 322G\\
&CMID (TGRS'2023) \cite{muhtar2023cmid}& Swin-B& $800^{2}$ & 75.11   &105M & 322G\\
&SkySense (CVPR'2024) \cite{guo2024skysense} & Swin-H& $800^{2}$ & 78.73  & \textgreater 674M & \textgreater 1679G\\
&\cellcolor{gray!10} RS-vHeat (Ours)& \cellcolor{gray!10} vHeat-B + SCEs & \cellcolor{gray!10} $800^{2}$ & \cellcolor{gray!10} \textbf{82.30} & \cellcolor{gray!10} 128M & \cellcolor{gray!10}\textbf{266G}\\
\bottomrule
\end{tabular}
\label{tab:dior}
}
\vspace{-0.7em}
\end{table*}

\begin{table*}[htbp]
\centering
\caption{
Comparison of OA, parameters, and FLOPs on AID and NWPU-RESISC45 with other RSFMs.}
\resizebox{1.0\linewidth}{!}{
\begin{tabular}{l|r|c|cc|cc|c|c|c}
\toprule
\multirow{2}{*}{ Backbone Type} & \multicolumn{1}{r|}{\multirow{2}{*}{ Method}} & \multirow{2}{*}{ Backbone} & \multicolumn{2}{c|}{AID} & \multicolumn{2}{c|}{ NWPU-RESISC45} & \multirow{2}{*}{ Image Size} & \multicolumn{1}{r|}{\multirow{2}{*}{ Params}} & \multirow{2}{*}{ FLOPs $\downarrow$} \\ \cline{4-7}

                               & \multicolumn{1}{r|}{}                        &                          &  TR=20\% &  TR=50\% &  TR=10\% &  TR=20\%              &                             & \multicolumn{1}{r|}{}                        &                        \\ \midrule

\multicolumn{1}{c|}{\multirow{5}{*}{CNN-Based}}
&GASSL (ICCV’2021) \cite{ayush2021geography} & ResNet-50  & 93.55 & 95.92&90.86& 93.06 & $1024^{2}$ & 24M & 87G\\
&SeCo (ICCV'2021) \cite{manas2021seasonal} & ResNet-50  & 93.47 & 95.99& 89.64&92.91  & $1024^{2}$ & 24M & 87G\\
&CACo (CVPR’2023) \cite{mall2023change} & ResNet-18   & 90.88&95.05 &88.28&91.94 & $1024^{2}$ & 11M &38G\\
&TOV (JSTARS’2023) \cite{tao2023tov}& ResNet-50  & 95.16& 97.09 & 90.97& 93.79 & $1024^{2}$ & 24M & 87G\\
&SSL4EO (GRSM'2023) \cite{wang2023ssl4eo} & ResNet-50  & 91.06&94.74 &87.60&91.27& $1024^{2}$ & 24M &87G\\
& \cellcolor{gray!10}RS-vHeat (Ours) & \cellcolor{gray!10}vHeat-B + SCEs & \cellcolor{gray!10}\textbf{96.81} & \cellcolor{gray!10}\textbf{97.58} &\cellcolor{gray!10} \textbf{92.01}  & \cellcolor{gray!10}\textbf{95.66} &\cellcolor{gray!10} $1024^{2}$ &\cellcolor{gray!10} 150M &\cellcolor{gray!10}340G \\
\midrule

\multicolumn{1}{c|}{\multirow{5}{*}{ViT-Based}}
&RVSA (TGRS'2022) \cite{wang2022advancing} & ViT-B + RVSA  &  \textbf{97.03}&\textbf{98.50} &\textbf{93.93}&\textbf{95.69} & $1024^{2}$ & 89M & 460G\\
&SatMAE (NeurIPS'2022)\cite{cong2022satmae}& ViT-L  &95.02&96.94&91.72&94.10 & $1024^{2}$ & 310M  & 2070G \\
&Scale-MAE (ICCV'2023) \cite{reed2023scale}& ViT-L    & 96.44&97.58 &92.63&95.04  & $1024^{2}$ & 310M  & 2070G  \\
& \cellcolor{gray!10} RS-vHeat (Ours) & \cellcolor{gray!10}vHeat-B + SCEs & \cellcolor{gray!10}96.81 &\cellcolor{gray!10} 97.58 &\cellcolor{gray!10} 92.01 & \cellcolor{gray!10}95.66 & \cellcolor{gray!10}$1024^{2}$ & \cellcolor{gray!10}150M &\cellcolor{gray!10}\textbf{340G}  \\
\midrule

\multicolumn{1}{c|}{\multirow{4}{*}{Swin-Based}}
&RingMo (TGRS'2022) \cite{sun2022ringmo}  & Swin-B & 96.90& 98.34 & 94.25& 95.67& $1024^{2}$ & 87M & 450G \\
&SatLas (ICCV'2023) \cite{bastani2023satlaspretrain}& Swin-B & 94.96&97.38&92.16&94.70& $1024^{2}$ & 87M & 450G \\
&GFM (ICCV'2023) \cite{mendieta2023towards}& Swin-B &  95.47&97.09 &92.73&94.64& $1024^{2}$ & 87M & 450G\\
&CMID (TGRS'2023) \cite{muhtar2023cmid}& Swin-B  &96.11&97.79 &94.05&95.53 & $1024^{2}$ & 87M & 450G \\
&SkySense (CVPR'2024) \cite{guo2024skysense} & Swin-H &\textbf{97.68} & \textbf{98.60} &\textbf{94.85 }&\textbf{96.32} & $1024^{2}$ & \textgreater  660M & \textgreater 2760G\\
& \cellcolor{gray!10}RS-vHeat (Ours) &\cellcolor{gray!10} vHeat-B + SCEs &\cellcolor{gray!10} 96.81 & \cellcolor{gray!10}97.58 & \cellcolor{gray!10}92.01 &\cellcolor{gray!10} 95.66 &\cellcolor{gray!10} $1024^{2}$ &\cellcolor{gray!10} 150M &\cellcolor{gray!10}\textbf{340G} \\
\bottomrule
\end{tabular}
}
\label{tab:classification}
\vspace{-1.5em}
\end{table*}

\begin{table}[htbp]
\centering
\setlength{\tabcolsep}{6pt}  
\caption{ Comparison of mAP\textsubscript{50} and mAP\textsubscript{75} on SAR-AIRcraft-1.0 with other specialized models.}
\resizebox{0.5\linewidth}{!}{
\begin{tabular}{r|c|c|c} 
\toprule
 Method &  Publication &  mAP\textsubscript{50} $\uparrow$ &  mAP\textsubscript{75} $\uparrow$  \\
\midrule
 Faster R-CNN \cite{ren2016faster} & TPAMI'2016 & 76.1 & 62.2\\
Cascade R-CNN \cite{cai2018cascade} & CVPR'2018 &75.7 & 58.9\\
RepPoints \cite{yang2019reppoints} & ICCV'2019 &72.6 & 53.3\\
SKG-Net \cite{fu2021scattering} & JSTARS'2021 &70.7  & 46.4\\
SA-Net \cite{zhirui2023sar} & RADARS'2023 &77.7 & 62.8\\

\cmidrule(lr){1-4}
\cellcolor{gray!10}RS-vHeat (Ours) &\cellcolor{gray!10}- & \cellcolor{gray!10}\textbf{87.1} &\cellcolor{gray!10} \textbf{67.4} \\
\bottomrule
\end{tabular}}
\label{tab:object_detection_SAR-AIRcraft}
\vspace{-1.2em}
\end{table}

\noindent\textbf{Change Detection.} We train and test on the LEVIR-CD dataset \cite{rs12101662}, employing the BIT architecture \cite{chen2021remote} with cross-entropy loss. RS-vHeat demonstrates superior adaptability, achieving an F1 score of 93.48\%, outperforming existing methods and surpassing SkySense \cite{guo2024skysense} by 0.9 points, as shown in \cref{tab:change_detection}. Despite a 256-pixel input size, RS-vHeat maintains FLOPs comparable to the Swin-B-based baseline while delivering higher precision. 

\begin{table}[htpb]
\centering
\setlength{\tabcolsep}{6pt}  
\caption{Comparison of mAP, parameters, and FLOPs on FAIR1M-2.0 with other RSFMs.}
\resizebox{0.5\linewidth}{!}{
\begin{tabular}{r|c|c|c|c}
\toprule
 Method &  Backbone &  mAP $\uparrow$  &  Params &  FLOPs $\downarrow$  \\
\midrule
CACo \cite{mall2023change} & ResNet-18 & 47.83 & 42M & 64G\\
GASSL \cite{ayush2021geography} & ResNet-50 & 48.15 & 55M & 77G \\
TOV \cite{tao2023tov}  & ResNet-50 &49.62  & 55M & 77G \\
SSL4EO \cite{wang2023ssl4eo} & ResNet-50 &  49.37  & 55M & 77G\\
RVSA \cite{wang2022advancing} & ViT-B + RVSA & 47.04  & 126M  & 160G\\
SatMAE \cite{cong2022satmae} & ViT-L &  46.55  & 336M & 241G\\
Scale-MAE \cite{reed2023scale} & ViT-L &  48.31  & 336M & 241G\\ 
SatLas \cite{bastani2023satlaspretrain} & Swin-B & 46.19 & 119M  & 167G\\
GFM \cite{mendieta2023towards} & Swin-B &  49.69 & 119M  & 167G\\
RingMo \cite{sun2022ringmo} & Swin-B & 46.21 & 119M  & 167G\\
CMID \cite{muhtar2023cmid} & Swin-B &  50.58  & 119M  & 167G \\
SkySense \cite{guo2024skysense} & Swin-H &  \textbf{54.57}  & \textgreater 688M  & \textgreater 900G\\
\midrule
\cellcolor{gray!10} RS-vHeat & \cellcolor{gray!10} vHeat-B + SCEs  &  \cellcolor{gray!10}48.29  &  \cellcolor{gray!10}130M  & \cellcolor{gray!10}137G  \\
\bottomrule
\end{tabular}
} 
\label{tab:FAIR1M-2.0}
\vspace{-1.2em}
\end{table}

\begin{table}[htb]
\centering
\setlength{\tabcolsep}{6pt}   
\caption{Comparison of F1 on LEVIR-CD with other RSFMs.}
\resizebox{0.5\linewidth}{!}{
\begin{tabular}{r|c|c|c|c}
\toprule
 Method &  Backbone &  F1 $\uparrow$ &  Params  &  FLOPs $\downarrow$  \\
\midrule
CACo  \cite{mall2023change} & ResNet-18 & 81.04 & 12M &11G\\
GASSL \cite{ayush2021geography} & ResNet-50  & 78.19 & 27M &25G\\
SeCo  \cite{manas2021seasonal} & ResNet-50  & 90.14 & 27M &25G\\
SSL4EO \cite{wang2023ssl4eo} & ResNet-50 & 89.05  & 27M &25G\\
RVSA \cite{wang2022advancing} & ViT-B + RVSA & 90.86 & 94M & 57G\\
SatMAE \cite{cong2022satmae}  & ViT-L &  87.65 & 304M &162G\\
Scale-MAE  \cite{reed2023scale} & ViT-L  & 92.07 & 304M &162G \\ 
SatLas \cite{bastani2023satlaspretrain} & Swin-B  & 90.62 & 88M &45G\\
GFM \cite{mendieta2023towards}  & Swin-B & 91.73 & 88M &45G\\
RingMo \cite{sun2022ringmo}  & Swin-B & 91.86 & 88M &45G\\
CMID \cite{muhtar2023cmid} & Swin-B  & 91.72 & 88M &45G\\
SkySense \cite{guo2024skysense} & Swin-H  & 92.58 & \textgreater 656M  &  \textgreater 307G\\
\midrule
\cellcolor{gray!10} RS-vHeat (Ours) &\cellcolor{gray!10} vHeat-B + SCEs & \cellcolor{gray!10}\textbf{93.48} & \cellcolor{gray!10}93M & \cellcolor{gray!10}46G\\
\bottomrule
\end{tabular}
\label{tab:change_detection}
}
\end{table}

\noindent\textbf{Image Classification.} We validate our model on two benchmark datasets (AID \cite{xia2017aid}, NWPU-RESISC45 \cite{cheng2017remote}) by attaching a classification head designed, and employ cross-entropy loss for computation. As shown in  \cref{tab:classification}, RS-vHeat surpasses CNN-based models like CACo \cite{mall2023change} in classification accuracy. Compared to ViT- and Swin-based models, RS-vHeat demonstrates superior computational efficiency while also achieving competitive accuracy. 

\subsection{Ablations}
To validate the effectiveness of the features learned through pre-training based on heat conduction theory, we conduct ablation studies on key components.

\begin{table*}[htbp]
\caption{Results of RS-vHeat under various constraints of self-supervised learning strategies.}
\centering
\setlength{\tabcolsep}{12pt}  
\resizebox{1.0\textwidth}{!}{
\begin{tabular}{c|ccc|cc|ccc}
\toprule
\multirow{3}{*}{\#} & \multicolumn{3}{c|}{\multirow{2}{*}{Loss}} & \multicolumn{2}{c|}{Object Detection}                        & \multicolumn{3}{c}{Semantic Segmentation}                                                                    \\ \cline{5-9} 
                    & \multicolumn{3}{c|}{}                      & \multicolumn{1}{c|}{DIOR (Optical)} & SAR-AIRcraft-1.0 (SAR) & \multicolumn{1}{c|}{iSAID (Optical)} & \multicolumn{1}{c|}{AIR-PolSAR-Seg (SAR)} & WHU-OPT-SAR (Optical+SAR) \\ \cline{2-9} 
                    & SDR           & FDR          & CL          & \multicolumn{1}{c|}{mAP\textsubscript{50} $\uparrow$}        & mAP\textsubscript{50} $\uparrow$                & \multicolumn{1}{c|}{mIoU $\uparrow$}          & \multicolumn{1}{c|}{mIoU $\uparrow$}               & OA $\uparrow$                      \\ \midrule
(a)                 & \Checkmark      & \ding{55}      & \ding{55}            & \multicolumn{1}{c|}{78.2}           & 84.6                   & \multicolumn{1}{c|}{66.1}            & \multicolumn{1}{c|}{54.7}                 & 80.5                      \\
(b)                  & \Checkmark   & \Checkmark   & \ding{55}          & \multicolumn{1}{c|}{79.5}           & 86.4                   & \multicolumn{1}{c|}{67.2}            & \multicolumn{1}{c|}{55.1}                 & 82.3                      \\
(c) & \Checkmark & \Checkmark & \Checkmark          & \multicolumn{1}{c|}{\textbf{82.3}}           & \textbf{87.1}                  & \multicolumn{1}{c|}{\textbf{68.7}}            & \multicolumn{1}{c|}{\textbf{57.5}}                 & \textbf{83.9}                      \\ \bottomrule
\end{tabular}
}
\label{tab:loss_comparison}
\end{table*}

\begin{figure*}[!htbp]
    \centering
    \includegraphics[width=\textwidth]{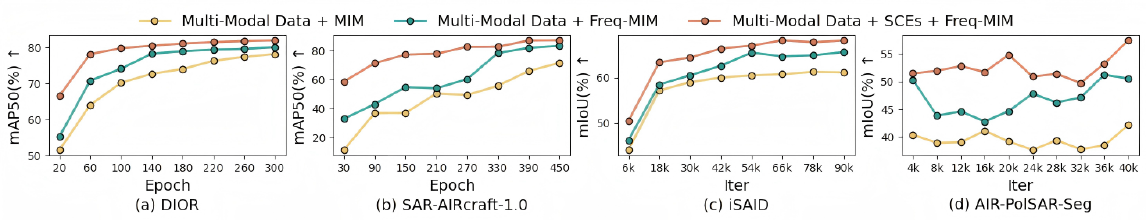}
    \caption{ Accuracy curves of various methods across modalities. (a) optical object detection on DIOR, (b) SAR object detection on SAR-AIRcraft-1.0,  (c) optical semantic segmentation on iSAID, and (d) SAR semantic segmentation on AIR-PolSAR-Seg.}
\label{fig:aaa_Effectiveness_of_Heat}
\end{figure*}

\noindent\textbf{Effectiveness of Masked Training.} We visualize the accuracy curves of the visual encoder before and after introducing new structures across downstream tasks, as shown in \cref{fig:aaa_Effectiveness_of_Heat}. All three networks are pre-trained on the same multi-modal RS data. In (a), we use vHeat-B with the SimMIM \cite{xie2022simmim} pixel mask training scheme; in (b), we add frequency domain masking and multi-domain reconstruction; and in (c), we integrate the innovative RS-vHeat structure, which includes SCEs. For optical tasks, RS-vHeat shows faster accuracy improvement, stabilizing at a higher ceiling. It surpasses 80\% mAP\textsubscript{50} on the DIOR dataset by the 120th epoch. For SAR tasks, while there are fluctuations, RS-vHeat demonstrates better adaptability and prediction accuracy. Overall, the improvements in pre-training and SCEs lead to advanced feature extraction performance.

\noindent\textbf{Reconstruction Learning Strategy.}
\cref{tab:loss_comparison} demonstrates the effectiveness of different loss components—spatial domain reconstruction loss (SDR), frequency domain reconstruction loss (FDR), and contrastive loss (CL). When only SDR is applied (row a), performance drops slightly on both optical and SAR datasets. Adding FDR (row b) improves performance by leveraging frequency domain information. Introducing CL (row c) further enhances performance by enforcing similarity constraints in the multi-modal feature space, improving feature learning. This leads to an 83.9\% OA on WHU-OPT-SAR, surpassing rows (a) and (b) by 3.4\% and 1.6\%, respectively. These results highlight the importance of combining these three losses for improved predictions across tasks and modalities.

\section{Conclusion}
In this work, we introduce the concept of heat conduction into RS tasks for the first time and establish a multi-modal RSFM, RS-vHeat. By employing a self-supervised learning strategy that integrates frequency domain masking and multi-domain reconstruction, along with a heat conduction operator incorporating spatial correction embeddings, we propose an approach that balances computational complexity and global receptive field coverage in RS. Our method captures global details from both spatial and frequency domains, significantly reducing the issue of small object omission, and achieves consistent multi-modal feature representation by mapping images into a high-dimensional thermal space. In future work, we plan to propose new solutions that offer fresh insights and methods for tackling visual modeling challenges across various industries.

\bibliographystyle{unsrtnat}
\bibliography{references}  






\clearpage


\begin{figure*}[!htbp]
    \centering
    \includegraphics[width=\linewidth]{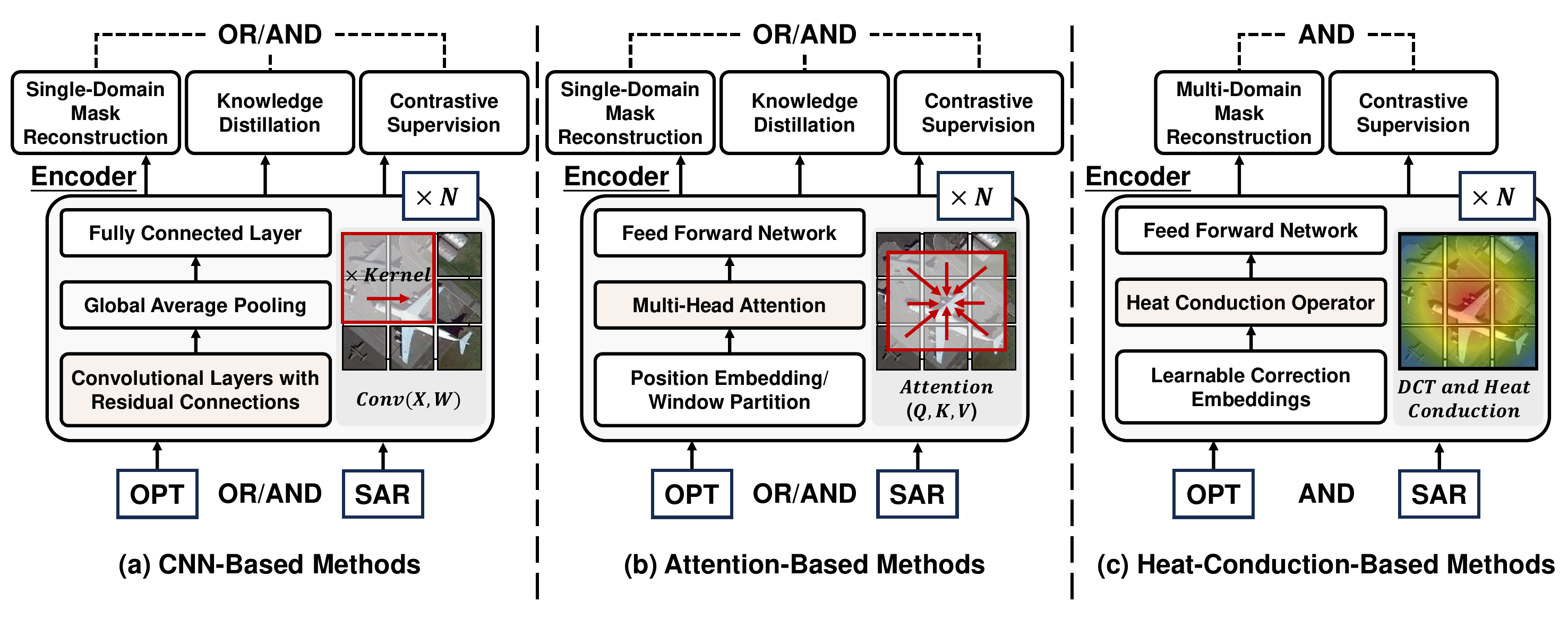}
\caption{Comparison of the self-supervised training scheme for the heat-conduction-based RSFM with other methods. (a) CNN-Based methods \cite{li2021geographical}, \cite{manas2021seasonal}, \cite{mall2023change}, \cite{ayush2021geography}. (b) Attention-Based methods \cite{cong2022satmae}, \cite{wang2022advancing}, \cite{tao2023tov}, \cite{reed2023scale}, \cite{mendieta2023towards}, \cite{bastani2023satlaspretrain}. (c) Heat-Conduction-Based method (ours). In our visual encoder, the heat conduction operator is employed to replace the residual blocks in CNN-based networks, and the attention layers in attention-based networks. For optical (OPT) and SAR inputs, the dual constraints of multi-domain mask reconstruction and distance metrics for multi-modal feature representations provide self-supervised signals during the pre-training process. This approach transforms the visual semantic propagation into a process of thermal diffusion within a thermal space, guided by the scene and object characteristics, dynamically extracting global information across the entire image. }
\label{fig:aaa_compare_architecture}
\end{figure*} 

\section*{A. Comparison details among RSFMs with different backbone networks}

Previous research on RSFMs primarily utilized existing visual encoders to extract deep features, integrating various self-supervised learning strategies with decoder structures, and pre-training on large-scale RS datasets. Visual encoders, as the core components of these models, are generally divided into two categories in recent research: 1) CNN-based methods \cite{li2021geographical}, \cite{manas2021seasonal}, \cite{mall2023change}, \cite{ayush2021geography}, as shown in \cref{fig:aaa_compare_architecture} (a). These models typically adopt the ResNet18/50 framework \cite{he2016deep}, with the residual module serving as the key learning structure. These approaches extract rich information from RS data through pixel masking reconstruction, expert geographical knowledge supervision or contrastive learning signals.  2) Attention-based methods, such as \cite{cong2022satmae}, \cite{wang2022advancing}, \cite{tao2023tov}, \cite{reed2023scale}, \cite{mendieta2023towards} and \cite{bastani2023satlaspretrain}, as illustrated in \cref{fig:aaa_compare_architecture} (b). These models primarily utilize the ViT \cite{dosovitskiy2020image} and Swin Transformers \cite{liu2021swin} as visual encoders, where the fundamental modules rely on attention mechanisms \cite{vaswani2017attention} and feed forward networks (FFNs) to model global dependencies. Pre-training is typically conducted through masked reconstruction, knowledge distillation or contrastive signals to enhance the robustness of the model representations. 

In summary, current RSFMs typically employ CNN-based or attention-based methods as visual encoders, innovating in learning and training strategies to enhance model performance. As shown in \cref{fig:aaa_compare_architecture} (c), RS-vHeat employs a heat-conduction-based visual encoder, with the heat conduction operator serving as the core computational module. During self-supervised learning, it applies frequency-domain and spatial-domain masking reconstruction constraints, along with an additional contrastive loss, which differentiates it significantly from existing RSFMs.

\section*{B. Preliminary of heat conduction}

Inspired by the physical principle of heat conduction, vHeat \cite{wang2024vheat} considers a region as a two-dimensional region $ D \in \mathbb{R}^2$. Then, for each point $(x, y)$ in the region, its temperature is $u(x, y, t)$ at time $t$, and the initial condition is $t=0$. The heat conduction propagation on this region can be expressed:
\begin{equation}
\frac{\partial u}{\partial t} = k(\frac{\partial^2 u}{\partial x^2} + \frac{\partial^2 u}{\partial y^2})
\label{heat_conduction_propagation}
\end{equation}
 where $k$ represents the thermal diffusivity. We denote the Fourier Transform and its inverse using the symbols $\mathcal{F}$ and $\mathcal{F}^{-1}$, respectively. After taking the Fourier Transform on both sides of the equals sign in 
 \cref{heat_conduction_propagation}, we formulate the calculation of physical heat equation as:
\begin{equation}
 \mathcal{F}(\frac{\partial u}{\partial t})= k\mathcal{F}(\frac{\partial^2 u}{\partial x^2} + \frac{\partial^2 u}{\partial y^2})
\label{Fourier_conduction_propagation}
\end{equation}

We represent the result of the Fourier transform of $u(x, y, t)$ as follows:
\begin{equation}
\tilde{u}(\omega_x,\omega_y,t) := \mathcal{F}(u(x, y, t))
\end{equation}

The left and right of \cref{Fourier_conduction_propagation} can be reformulated as
\begin{equation}
 \mathcal{F}(\frac{\partial u}{\partial t})= \frac{\partial \tilde{u}(\omega_x,\omega_y,t)}{\partial t}
\end{equation}
\begin{equation}
 \mathcal{F}(\frac{\partial^2 u}{\partial x^2} + \frac{\partial^2 u}{\partial y^2}) = -(\omega^2_x +\omega^2_y)\tilde{u}(\omega_x,\omega_y,t)
\end{equation}

Furthermore, the \cref{Fourier_conduction_propagation} is expressed as an ordinary differential equation in the frequency domain:
\begin{equation}
 \frac{d\tilde{u}(\omega_x,\omega_y,t)}{dt} = -k(\omega^2_x +\omega^2_y)\tilde{u}(\omega_x,\omega_y,t)
\label{ode}
\end{equation}

To solve $\tilde{u}(\omega_x,\omega_y,t)$ in \cref{ode}, we use $\tilde{f}(\omega_x,\omega_y)$ to represent the Fourier Transform of $f(x, y)$, and we can get the following result under the initial condition of $\tilde{u}(\omega_x,\omega_y,t)|_{t=0}$:
\begin{equation}
\tilde{u}(\omega_x,\omega_y,t) = \tilde{f}(\omega_x,\omega_y)e^{-k(\omega^2_x +\omega^2_y)t}
\end{equation}

Finally, the values in the frequency domain are converted back to the space domain by inverse Fourier Transform, and we get the general solution of heat equation in the spatial domain expressed as follows:
\begin{equation}
u(x,y,t) = \mathcal{F}^{-1}(\tilde{f}(\omega_x,\omega_y)e^{-k(\omega^2_x +\omega^2_y)t}) = \frac{1}{4\pi^2}\int_{\tilde{D}}\tilde{f}(\omega_x,\omega_y)e^{-k(\omega^2_x +\omega^2_y)t}e^{i(\omega_xx +\omega_yy)}d\omega_xd\omega_y
\label{inverse}
\end{equation}

\section*{C. Implementation details of the masking strategy}

Given the multi-modal input (optical and SAR), denoted as \( I(x,y,c)=\{I_{o}, I_{s}\}, I_{o} \in \mathbb{R}^{\mathit{H \times W \times} 3}, I_{s} \in \mathbb{R}^{\mathit{H \times W \times} 1} \), the process begins by applying the DCT along each image dimension $c = 1, \dots, C $, extracting 2D planes from the spatial domain \( I(x, y) \) and converting them into its frequency representation $\tilde{I}(u,v)$.  This transformation concentrates low-frequency information in the top-left corner of the frequency spectrum:
\begin{equation}
\tilde{I}(u,v) = \frac{2}{\sqrt{MN}} \sum_{x=0}^{M-1} \sum_{y=0}^{N-1} I(x,y) \cos \frac{(2x+1)u\pi}{2M} \cos \frac{(2y+1)v\pi}{2N} 
\end{equation}
where $M$ and $N$ denote the width and height of the input image, respectively.

To address signals across different frequency ranges, we apply a sector mask to the transformed image. Centered at the top-left, this mask separates the image into distinct high-frequency $\tilde{I}^{high}(u,v)$ and low-frequency $\tilde{I}^{low}(u,v)$  regions:
\begin{equation}
\tilde{I}^{low}(u,v), \tilde{I}^{high}(u,v) = \tilde{M} \odot \tilde{I}(u,v)
\end{equation}
The binary mask $\tilde{M}$, sized $(M \times N)$, is applied to each dimension $c$ using the operator $\odot$. Each element of $\tilde{M}$ takes a value of either 0 or 1.

After applying the mask, we perform the IDCT to convert the processed frequency representation back to its spatial representation along each dimension:
\begin{equation}
I^{low}(x,y) = \sum_{u=0}^{M-1} \sum_{v=0}^{N-1} \frac{2}{\sqrt{MN}}  \tilde{I}^{low}(u,v) \cos \frac{(2x+1)u\pi}{2M} \cos \frac{(2y+1)v\pi}{2N}
\end{equation}

\begin{equation}
I^{high}(x,y) = \sum_{u=0}^{M-1} \sum_{v=0}^{N-1} \frac{2}{\sqrt{MN}}  \tilde{I}^{high}(u,v) \cos \frac{(2x+1)u\pi}{2M} \cos \frac{(2y+1)v\pi}{2N}
\end{equation}
Where \(I^{low}(x,y)\) and \(I^{high}(x,y)\) denote the low- and high-frequency representation that are converted back to their spatial domain after applying the mask. The results are then concatenated to restore the original dimensionality. 

\section*{D. Configuration and visualization results of fownstream task datasets}
RS-vHeat is trained on 10 datasets across 4 downstream tasks. In this section, we provide detailed information about the datasets and experimental configurations.

\begin{figure*}[htbp]
    \centering
    \includegraphics[width=1.0\linewidth]{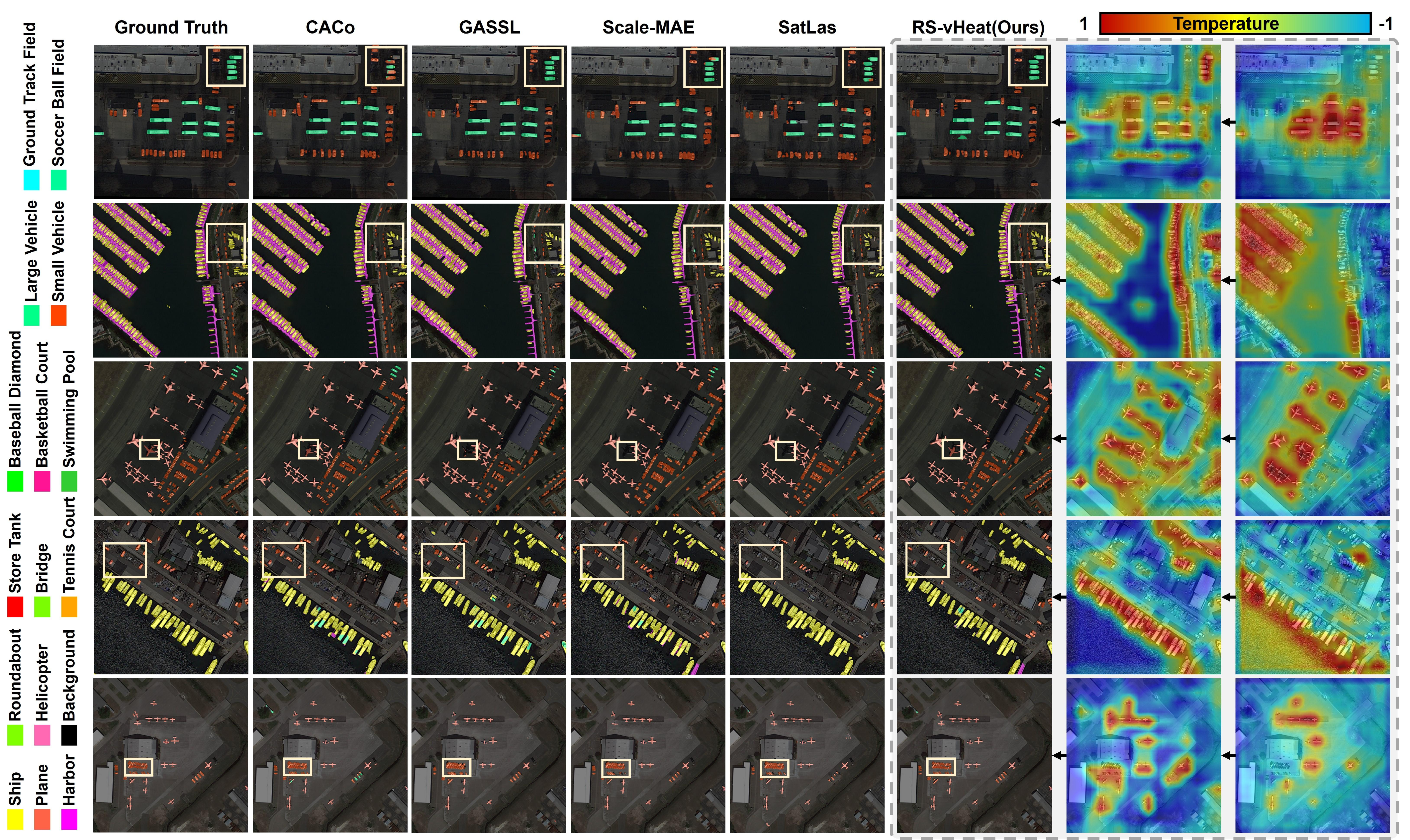}
\caption{The qualitative results of RS-vHeat and several representative RSFMs on the iSAID dataset. Each column from left to right represents: ground truth, CACo (ResNet-18), GASSL (ResNet-50), Scale-MAE (ViT-L), Satlas (Swin-B), and the results from our model, RS-vHeat. The last two columns on the right visualize the output variations of RS-vHeat across the final two stages.}
\label{fig:aaa_isaid_vis}
\end{figure*}

\begin{figure*}[htbp]
    \centering
    \includegraphics[width=\linewidth]{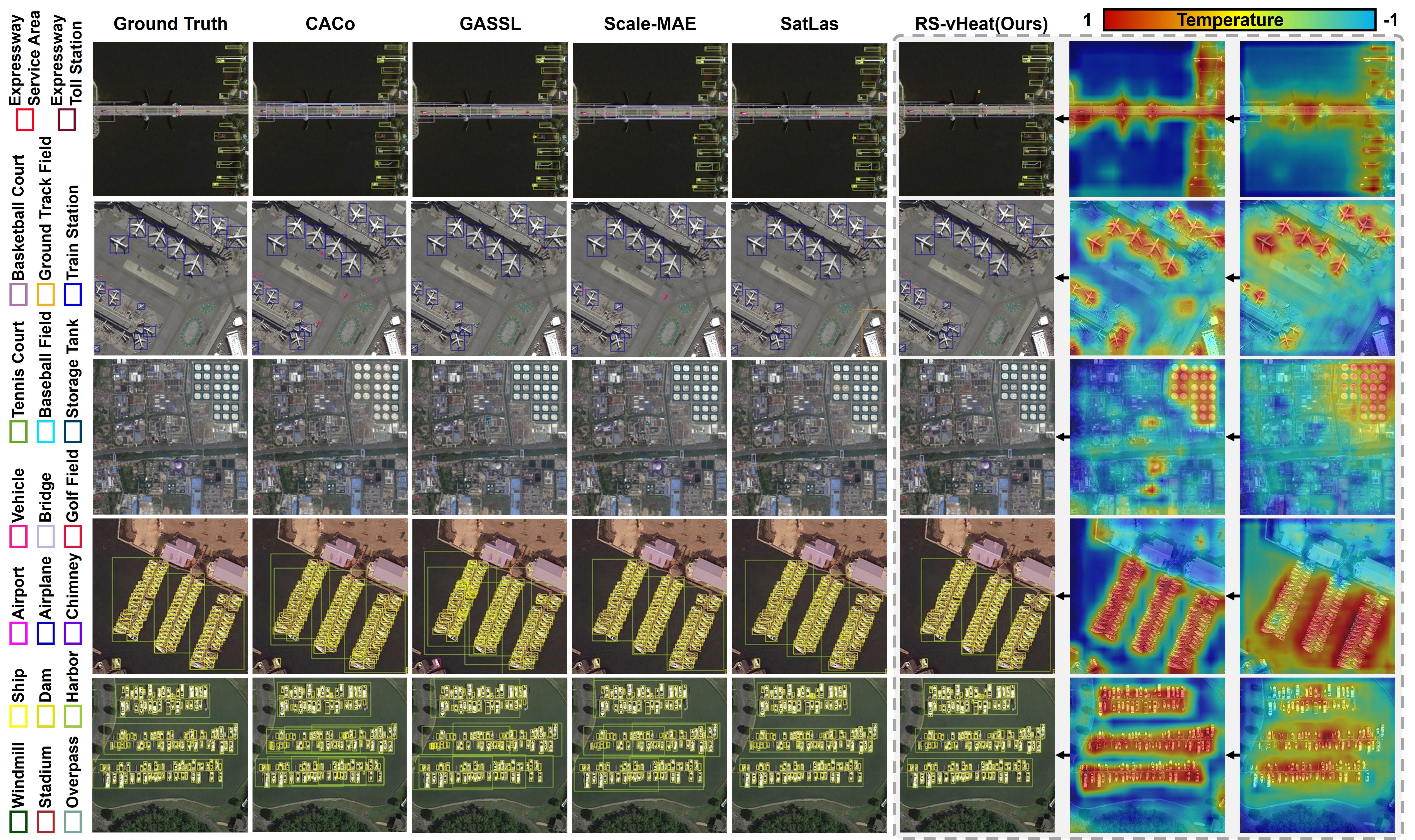}
\caption{The qualitative results of RS-vHeat and several representative RSFMs on the DIOR dataset. Each column from left to right represents: ground truth, CACo (ResNet-18), GASSL (ResNet-50), Scale-MAE (ViT-L), Satlas (Swin-B), and the results from our model, RS-vHeat. The last two columns on the right visualize the output variations of RS-vHeat across the final two stages.} 
\label{fig:aaa_dior_vis}
\end{figure*}

\begin{table*}[htbp]
\centering
\setlength{\tabcolsep}{12pt}  
\caption{ Comparison of AP\textsubscript{50} for each category, mAP\textsubscript{50} and mAP\textsubscript{75} on SAR-AIRcraft-1.0 with other specialized models.}
\resizebox{1.0\textwidth}{!}{
\begin{tabular}{r|c|c|c|c|c|c|c|c|c|c} 
\toprule
 Method &  Publication &  A330 &  A320/A321 &  A220 &  ARJ21 &  Boeing737 &  Boeing787 &  Other &  mAP\textsubscript{50} $\uparrow$ &  mAP\textsubscript{75} $\uparrow$  \\
\midrule
 Faster R-CNN \cite{ren2016faster} & TPAMI'2016 & 85.0 & 97.2 & 78.5 & 74.0 & 55.1 & 72.9 & 70.1 & 76.1 & 62.2\\
Cascade R-CNN \cite{cai2018cascade} & CVPR'2018 &87.4 &97.5 &74.0 &78.0 &54.5 &68.3 &69.1 &75.7 & 58.9\\
RepPoints \cite{yang2019reppoints} & ICCV'2019 &89.8 & \textbf{97.9} &71.4 &73.0 &55.7 &51.8 &68.4 &72.6 & 53.3\\
SKG-Net \cite{fu2021scattering} & JSTARS'2021 &79.3 &78.2 &66.4 &65.0 &65.1 &69.6 &71.4 &70.7  & 46.4\\
SA-Net \cite{zhirui2023sar} & RADARS'2023 &88.6 &94.3 &80.3 &78.6 &59.7 &70.8 &71.3 &77.7 & 62.8\\
\cmidrule(lr){1-11}
\cellcolor{gray!10}RS-vHeat (Ours) &\cellcolor{gray!10}- &  \cellcolor{gray!10}\textbf{98.4} & \cellcolor{gray!10}\textbf{ 97.9} & \cellcolor{gray!10} \textbf{81.1}  &  \cellcolor{gray!10}\textbf{89.3} &\cellcolor{gray!10}  \textbf{82.0}  &\cellcolor{gray!10}  \textbf{79.8} &  \cellcolor{gray!10}\textbf{81.1} &\cellcolor{gray!10}  \textbf{87.1}  & \cellcolor{gray!10} \textbf{67.4} \\
\bottomrule
\end{tabular}}
\label{tab:object_detection_SAR-AIRcraft_fine}
\vspace{-1.0em}
\end{table*}

\noindent\textbf{D.1. Single- and multi-modal semantic segmentation}

We utilize RS-vHeat as the visual encoder and implemented UPerNet \cite{Xiao_2018_ECCV} with cross-entropy loss for the output head. Additionally, we employ the AdamW optimizer with a learning rate of 6e-5 and conduct a warm-up of 1500 iterations.

\noindent\textbf{Dataset.}
We evaluated our model on three single-modal datasets and one multi-modal dataset:

1) The Potsdam dataset \cite{sherrah2016fully} comprises 38 images. This dataset is annotated with six classes, each having a resolution of $6000 \times 6000$ pixels. The input resolution is set to 512 pixels.

2) The iSAID dataset \cite{waqas2019isaid} comprises 2,806 images with varying resolutions, primarily focusing on urban environments. The dataset includes annotations for 15 different categories and we utilize an image size of 896 pixels as the input for the model.

3) The Air-PolSAR-Seg dataset \cite{wang2022air} focuses on polarimetric SAR images. It offers a region measuring \(9082 \times 9805\) pixels and includes 2,000 image patches, each sized $512 \times 512$. The dataset features pixel-wise annotations covering six categories. We adopt a size of 512 pixels for the image input.

4) The WHU-OPT-SAR dataset \cite{li2022mcanet} is a multi-modal segmentation dataset with a resolution of 5 meters. It includes optical and SAR data from the same region, categorized into seven classes. Each image has a size of $5556 \times 3704$  pixels. We uniformly cropped the multi-modal images to a pixel size of 256 for model input.

\noindent\textbf{Metric.} Following the configurations of RingMo \cite{sun2022ringmo} and SkySense \cite{guo2024skysense}, we evaluate the mean Intersection over Union (mIoU) on the iSAID dataset and test the mean F1 score (mF1) on the Potsdam dataset. For the AIR-PolSAR-Seg dataset, we use three metrics: mIoU, Overall Accuracy (OA) and Average Accuracy (AA). We assess OA and User's Accuracy on the WHU-OPT-SAR dataset following the setup outlined in the corresponding paper.

\noindent\textbf{Additional Results.}  The \cref{fig:aaa_isaid_vis} displays the process visualizations and prediction results for the iSAID dataset, which display that the heat-conduction-based backbone exhibits adaptive characteristics when capturing features across different layers.

\noindent\textbf{D.2. Object Detection}

We conduct coarse- and fine-grained experiments on optical and SAR datasets to demonstrate the robustness of RS-vHeat. In the horizontal bounding boxes (HBB) task, we employ SGD as the optimizer, with a base learning rate set to 0.01. A warm-up phase of 3 epochs is conducted. YOLOX \cite{ge2021yolox} is used as the output head, and experiments are conducted using cross-entropy loss and IoU loss. In the oriented bounding box (OBB) task, we adjust the base learning rate to 1e-4. The warm-up phase consists of 500 iterations. Oriented RCNN \cite{xie2021oriented} is used as the output head, applying cross-entropy and Smooth \( \mathcal{L}_1 \) loss.

\noindent\textbf{Dataset}.
Our model is tested on three challenging object detection datasets:

1) FAIR1M \cite{sun2022fair1m} is an optical fine-grained dataset with objects annotated using OBB, encompassing five major categories, further divided into 37 subcategories. The dataset contains over 40,000 images. Following the official split, we ultimately submitted the test results to the website to obtain accuracy measurements. We utilize an image size of 512 pixels as the input for the model.

2) SAR-AIRcraft-1.0 \cite{zhirui2023sar} is a HBB fine-grained SAR aircraft object detection dataset designed for challenging scenarios, totaling 4,368 images. It encompasses seven fine-grained categories. We adopt a size of 640 pixels for the image input.

3) DIOR \cite{li2020object} is an optical dataset that includes 20 categories. It comprises a total of 23,463 images and provides HBB annotations. We utilize an image size of 800 pixels as the input for the model.

\noindent\textbf{Metric}.
On the FAIR1M and DIOR dataset, we evaluate the mAP (Mean Average Precision). For the SAR-AIRcraft-1.0 dataset, we evaluate the AP\textsubscript{50} for each category, mAP\textsubscript{50} and mAP\textsubscript{75}. mAP\textsubscript{50} and mAP\textsubscript{75} represent the mAP at IoU thresholds of 0.5 and 0.75, respectively, with category-specific precision calculated at an IoU threshold of 0.5.

\noindent\textbf{Additional Results.} The visualization results of the DIOR dataset are shown in \cref{fig:aaa_dior_vis}. From the feature extraction process and results, RS-vHeat outperforms other RSFMs in terms of extracting dense RS objects. Additionally, we further refine the RS-vHeat extraction results for each class of the SAR-AIRcraft-1.0 dataset in \cref{tab:object_detection_SAR-AIRcraft_fine}, highlighting its enhanced capability in recognizing various aircraft types in SAR scenarios compared to specialized object detection models.

\noindent\textbf{D.3. Change Detection}

We employ RS-vHeat as the visual encoder, accommodating images before and after transformation. AdamW optimizer is used with a base learning rate of 0.002 and we train for 200 epochs. The BIT architecture \cite{chen2021remote} is utilized for subsequent image change analysis, with cross-entropy loss applied for the experiments.

\noindent\textbf{Dataset.}
We use the LEVIR-CD dataset to train and test:

1) The LEVIR-CD dataset \cite{rs12101662} consists of 637 image patch pairs obtained from Google Earth. Each patch has a size of $1024 \times 1024$ pixels. The dataset primary focus is on building-related changes, such as the emergence of new structures and the decline of existing ones. We utilize an image size of 256 pixels as the input.

\noindent\textbf{Metric.}
We use F1-score to evaluate change detection performance. F1-score is the harmonic mean of precision and recall, providing a balanced measure of performance.

\noindent\textbf{D.4. Image Classification}

We extend our model by attaching a classification head designed to handle the classification task and employ cross-entropy loss for computation. We utilize AdamW as the optimizer with a learning rate of 5e-4, training for 300 epochs.

\noindent\textbf{Dataset.} We validate our model on two benchmark datasets as described below.

1) The Aerial Image Dataset (AID) \cite{xia2017aid} consists of 30 categories, with each category containing approximately 220 to 420 images sized at $600 \times 600$ pixels, totaling 10,000 images. 

2) The NWPU-RESISC45 dataset \cite{cheng2017remote} is a RS image dataset comprising 45 categories, with a total of 31,500 images distributed across these categories. Each category consists of 700 images.

\noindent\textbf{Metric.}
We use OA to evaluate classification performance. We follows standard practices in the field \cite{guo2024skysense}, using $20\%$ and $ 50\% $ of the AID dataset as training sets, and $10\%$ and $20\%$ of the NWPU-RESISC45 dataset as training sets.

\end{document}